\documentclass[11pt,a4paper]{article}
\usepackage[hyperref]{acl2018}
\usepackage{times}
\usepackage{latexsym}
\usepackage{amsmath}
\usepackage{adjustbox}
\usepackage{graphicx}
\usepackage{url}
\usepackage{siunitx}

\aclfinalcopy 

\makeatletter
\def\blfootnote{\xdef\@thefnmark{}\@footnotetext}
\makeatother

\newcommand\nnfootnote[1]{%
  \begin{NoHyper}
  \renewcommand\thefootnote{}\footnote{#1}%
  \addtocounter{footnote}{-1}%
  \end{NoHyper}
}

\title{CAiRE\_HKUST at SemEval-2019 Task 3: Hierarchical Attention for Dialogue Emotion Classification}

\author{
Genta Indra Winata*, Andrea Madotto*, Zhaojiang Lin,  \\\textbf{Jamin Shin, Yan Xu, Peng Xu, Pascale Fung} \\
  Center for Artificial Intelligence Research (CAiRE)\\
  Department of Electronic and Computer Engineering\\
  The Hong Kong University of Science and Technology, Clear Water Bay, Hong Kong\\
  \texttt{\{giwinata,amadotto,zlinao\}@connect.ust.hk},\\ \texttt{\{jmshinaa,yxucb,pxuab\}@connect.ust.hk,pascale@ece.ust.hk}
}

\date{}

\begin{document}
\maketitle
\begin{abstract}
Detecting emotion from dialogue is a challenge that has not yet been extensively surveyed. One could consider the emotion of each dialogue turn to be independent, but in this paper, we introduce a hierarchical approach to classify emotion, hypothesizing that the current emotional state depends on previous latent emotions. We benchmark several feature-based classifiers using pre-trained word and emotion embeddings, state-of-the-art end-to-end neural network models, and Gaussian processes for automatic hyper-parameter search. In our experiments, hierarchical architectures consistently give significant improvements, and our best model achieves a 76.77\% F1-score on the test set.
\end{abstract}

\section{Introduction}
\nnfootnote{*Equal contribution.} Customer service can be challenging for both the givers and receivers of services, leading to emotions on both sides. 
Even human service-people who are trained to deal with such situations struggle to do so, partly because of their own emotions. Neither do automated systems succeed in such scenarios. \textit{What if we could teach machines how to react under these emotionally stressful situations of dealing with angry customers?}

This paper represents work on the SemEval 2019 shared task~\cite{SemEval2019Task3}, which aims to bring more research on teaching machines to be empathetic, specifically by contextual emotion detection in text. Given a textual dialogue with two turns of context, the system has to classify the emotion of the next utterance into one of the following emotion classes: Happy, Sad, Angry, or Others. The training dataset contains 15K records for emotion classes, and contains 15K records not belonging to any of the aforementioned emotion classes. 

The most naive first step would be to recognize emotion from a given flattened sequence, which has been researched extensively despite the very abstract nature of emotion \cite{socherRNTN, deepmoji, bcn, xu2018emo2vec, chatterjee2019understanding}. However, these \textit{flat} models do not work very well on dialogue data as we have to merely concatenate the turns and flatten the hierarchical information. Not only does the sequence get too long, but the hierarchy between sentences will also be destroyed \citep{hsu2018socialnlp, kim2018attnconvnet}. We believe that the natural flow of emotion exists in dialogue, and using such hierarchical information will allow us to predict the last utterance's emotion better.

Naturally, the next step is to be able to detect emotion with a hierarchical structure. To the best of our knowledge, this task of extracting emotional knowledge in a hierarchical setting has not yet been extensively explored in the literature. Therefore, in this paper, we investigate this problem in depth with several strong hierarchical baselines and by using a large variety of pre-trained word embeddings. 

\section{Methodology}

\begin{figure*}[!t]
    \centering
    \includegraphics[width=0.8\linewidth]{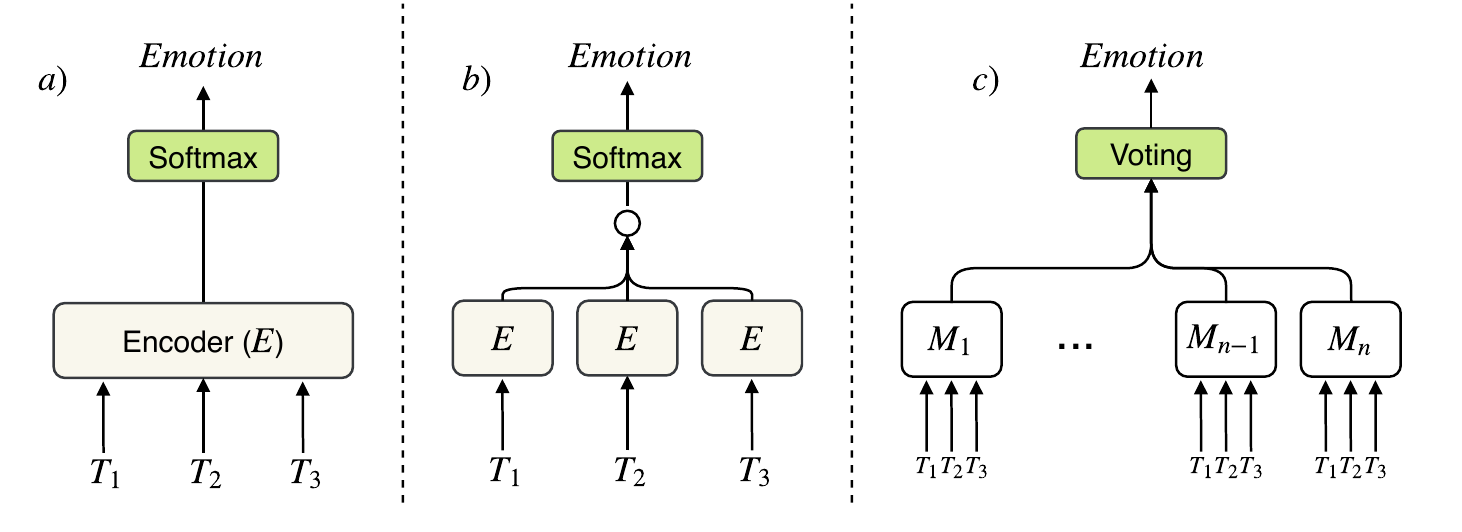}
    \caption{a) Flat model; b) Hierarchical model; c) Voting scheme }
    \label{fig:architecture}
\end{figure*}

In this task, we focus on two main approaches: 1) feature-based and 2) end-to-end. The former compares several well-known pre-trained embeddings, including \textit{GloVe} \cite{pennington2014glove}, \textit{ELMo} \cite{peters2018deep}, and \textit{BERT} \cite{bert}, as well as emotional embeddings. We combine these pre-trained features with a simple Logistic Regression (LR) and XGBoost \cite{xgboost} model as the classifier to compare their effectiveness. The latter approach is to train a model fully end-to-end with back-propagation. We mainly compare the performances of \textit{flat} models and \textit{hierarchical} models, which also take into account the sequential turn information of dialogues.

\subsection{Feature-based Approach}
The pre-trained feature-based approach can be subdivided into two categories: 1) word embeddings pre-trained only on semantic information, and 2) emotional embeddings that augment word embeddings with emotional or emoji information. We also examine the use of both categories.

\paragraph{Word Embeddings}
These include the standard pre-trained non-contextualized GloVe \cite{pennington2014glove}, the contextualized embeddings from the bidirectional long short term memory (biLSTM) language model ELMo \cite{peters2018deep}, and the more recent \textit{transformer} based embeddings from the bidirectional language model BERT \cite{bert}.

\paragraph{Emotional Embeddings} These refer to two types of features equipped with emotional knowledge. The first is a word-level emotional representation called Emo2Vec \cite{xu2018emo2vec}. It is trained with six different emotion-related tasks and has shown extraordinary performance over 18 different datasets. The second is a sentence-level emotional representation called DeepMoji \cite{felbo2017using}, trained with a biLSTM with an attention model to predict emojis from text on a 1,246 million tweet corpus. Finally, we use Emoji2Vec \cite{eisner2016emoji2vec} which directly maps emojis to continuous representations.

\subsection{End-to-End Approach}
We consider four main models for the end-to-end approach: fine-tuning ELMo \cite{peters2018deep},  fine-tuning BERT \cite{bert}, Long Short Term Memory (LSTM) \cite{hochreiter1997long}, and Universal Transformer (UTRS) \cite{universal_transformer}.\footnote{We also tested Transformer, but had an overfitting issue} In the latter model, we also run a Gaussian process for automatic hyper-parameter selection.

\paragraph{ELMo} This model from \citet{peters2018deep} is a deep contextualized embedding extracted from a pre-trained bidirectional language model that has shown state-of-the-art performance in several natural language processing (NLP) tasks. 


\paragraph{BERT} This is the state-of-the-art bidirectional pre-trained language model that has recently shown excellent performance in a wide range of NLP tasks. Here, we use $BERT_{BASE}$\footnote{We used a PyTorch implementation from https://github.com/huggingface/pytorch-pretrained-BERT} as our sentence encoder. However, the original model failed to capture the emoji features due to the fact that all the emoji tokens are missing in the vocab. Therefore, we concatenate each sentence representation from BERT with bag of words \textit{Emoji2Vec} \cite{eisner2016emoji2vec}. Then, a UTRS is used as a context encoder to encode the whole sequence.

\paragraph{LSTM and Universal Transformer} 
LSTM is the widely known model used almost ubiquitously in the literature, while UTRS is a recently published recurrent extension of the multi-head self-attention based model, Transformer from \cite{transformer}. Finally, for all models, we consider a hierarchical extension which considers the turn information as well. We add another instance of the same model to also encode sentence-level information on top of the word-level representations. We also apply word-level attention to select the important information words on each dialogue turn.


\begin{table}[!t]
\centering
\caption{The table shows the F1 score on LR and XGBoost.}
\label{results-feature-based}
\resizebox{0.37\textwidth}{!}{
\begin{tabular}{lcc}
\hline
\textbf{Feature(s)} & \textbf{Classifier} & \textbf{F1} \\ \hline \hline
DeepMoji & LR & 64.87 \\
ELMo & LR & 63.86 \\
GLoVe & LR & 55.11 \\
Emo2Vec & LR & 50.91 \\
BERT & LR & 44.51 \\
Emoji2Vec & LR & 30.45 \\ \hline
\textbf{ELMo + DeepMoji} & \textbf{LR} & \textbf{65.63} \\
ELMo + Emo2Vec & LR & 65.42 \\
Emoji2Vec + GLoVe & LR & 58.00 \\ \hline
\textbf{ELMo + DeepMoji} & \textbf{XGBoost} & \textbf{69.86} \\ \hline
\end{tabular}
}
\end{table}

\begin{table}[!t]
\centering
\caption{The table shows F1 score on flat and hierarchical end-to-end models. GP denotes as Gaussian process.}
\label{results-end2end}
\resizebox{0.47\textwidth}{!}{
\begin{tabular}{lcc}
\hline
\textbf{Model} & \multicolumn{1}{c}{\textbf{Flat}} & \multicolumn{1}{c}{\textbf{Hierarchical}} \\ \hline \hline
LSTM & 72.53 & 73.45 \\
\textbf{LSTM+GLoVe} & \textbf{73.95} & \textbf{75.64} \\
LSTM+GLoVe+Emo2Vec & 73.85 & 74.59 \\
UTRS & 72.41 & 74.06 \\ 
ELMo & 68.14 & 70.55 \\
BERT & 66.12 & 73.29 \\ \hline
\end{tabular}
}
\end{table}

\begin{table}[!t]
\centering
\caption{The table shows F1 score on different ensemble models. XGB denotes XGBoost with ELMo and DeepMoji features. ALL denotes all ensemble models.}
\label{results-ensemble}
\resizebox{0.46\textwidth}{!}{
\begin{tabular}{lcc}
\hline
\textbf{Model} &  \multicolumn{1}{c}{\textbf{F1}} \\ \hline \hline
Ensemble$_1$ \text{ (3 HLSTMs)} & 76.08 \\
Ensemble$_2$ \text{ (HBERT + HLSTM + HUTRS)} & 75.76 \\
Ensemble$_3$ \text{ (HBERT + 3 HLSTMs + HUTRS)} & 76.26 \\
Ensemble$_4$ \text{ (HBERT + 5 HLSTMs + HUTRS)} & 76.24 \\
Ensemble$_5$ \text{ (HBERT + 5 HLSTMs + HUTRS)} & 76.20 \\ \hline
\textbf{Ensemble}$_{final}$ \textbf{ (ALL + HLSTM + XGB) } & \textbf{76.77} \\ 
- Angry & 75.88\\
- Happy & 73.65\\
- Sad & 81.30\\ \hline
\end{tabular}
}
\end{table}

\section{Evaluation}
In this section, we present the evaluation metrics used in the experiment, followed by results on feature-based, end-to-end, and ensemble approaches and Gaussian process search.

\begin{figure}[!t]
    \centering
    \includegraphics[width=0.9\linewidth]{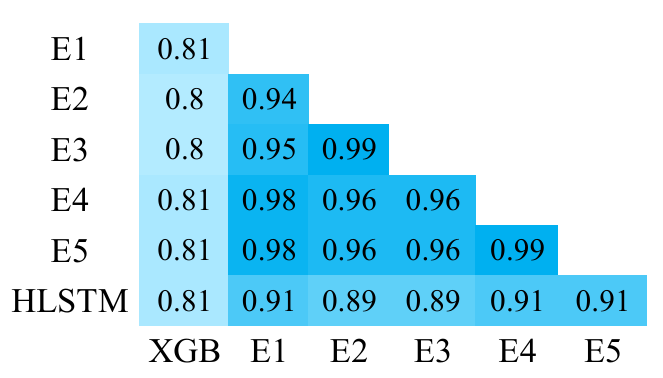}
    \caption{Pearson correlation matrix of a model to other models. E1--E5 denote ensemble models.}
    \label{fig:correlation}
\end{figure}

\subsection{Training Details}
\paragraph{Feature-Based} For the feature-based approach, we run LR and XGBoost on features using the Scikit-Learn toolkit \cite{scikit-learn} without any additional tuning. 

\paragraph{ELMo} For the flat model, we pre-train ELMo by only fine-tuning the scalar-mix weights, as suggested in~\citet{peters2018deep}. We extract a 1024-dimension bag-of-words representation for each turn and concatenate the three turns into a 3072-dimension vector which is passed to a multilayer perceptron (MLP). For the hierarchical model, we employ two methods: 1) run an LSTM model over each turn's representation 2) pre-extract all three layer weights (LSTM and CNN) and concatenate them into a 3072-dimension vector representation for each turn, which is then passed to an LSTM model. We report the results of the latter pre-extracted method as it performs better.

\paragraph{BERT} For the implementation details of $BERT_{BASE}$, we refer interested readers to \citet{bert}. Note that for hierarchical BERT, we use a six-layer UTRS as the context encoder.~Each layer of UTRS consists of a multi-head attention block with four heads, where the dimension of each head is set to be ten, and a convolution feed forward block with 50 filters. We use modified Adam optimizer from \citet{bert} to train our model. The initial learning rate and dropout are 5e-5 and 0.3 respectively.

\paragraph{LSTM and Universal Transformer} We train hierarchical LSTMs with hidden sizes of \{1000, 1500\} using different dropouts \{0.2,0.3,0.4,0.5\}. The best LSTMs (without additional features, with GLoVE, with GLoVE+Emo2Vec) reported in Figure \ref{results-end2end} have a hidden size of 1000 and dropout of 0.5, a hidden size of 1500 and dropout of 0.2, and a hidden size of 1000 and dropout of 0.4 respectively. Then, we train the UTRS using the best hyper-parameters found by the GP. It has a hidden size of 488 with a single hop and ten attention multi-heads. Noam \cite{transformer} is used as the learning rate decay.

\paragraph{Gaussian Processes} GP hyper-parameter search returns a set of hyper-parameters, both continuous and discrete, and it returns the validation set F1 score. We implement the GP model using an existing library called GPyOpt.\footnote{http://sheffieldml.github.io/GPyOpt/} We run a GP for 100 iterations using the Expected Improvement  \cite{jones1998efficient} acquisition function with 0.05 jitter as a starting point. We use a hierarchical universal transformer (HUTRS) as the base model since is the model with the most hyper-parameters to tune with a single split.

\subsection{Evaluation Metrics}
The task is evaluated with a micro F1 score for the three emotion classes, i.e.,~Happy, Sad and Angry, and by taking the harmonic mean of the precision and the recall. This scoring function has been provided by the challenge organizers \cite{SemEval2019Task3}.

\subsection{Voting Scheme}
For each model, we randomly shuffle and split the training set ten times and we apply a voting scheme to create a more robust prediction. We use a majority vote scheme to select the most often seen predictions, and in case of ties, we give the priority to \textit{Others}. This scheme is applied to all end-to-end models since it improved the validation set performance.

\subsection{Ensemble Models}
To further refine our predictions, we build ensembles of different models.~We create five ensemble models by combining the hierarchical version of BERT, LSTM, and UTRS. Finally, we gather two lesser-performing models, a hierarchical LSTM and the best feature-based model (XGBoost with ELMo and DeepMoji features), and we combine them with five ensemble predictions using majority voting to get our final prediction.~Finally, we show the Pearson correlation between models in Figure \ref{fig:correlation}.

\subsection{Experimental Results}
From Table \ref{results-feature-based}, we can see that the DeepMoji features outperforms all the other features by a large margin. Indeed, DeepMoji has been trained using a large emotion corpus, which is compatible with the current task. Emoji2Vec get a very low F1-score since it includes only emojis, and indeed, by adding GLoVe, a more general embedding, we achieve better performance. For the end-to-end approach, hierarchical biLSTM with GLoVe word embedding achieves the highest score with a 75.64\% F1-score. Our ensemble achieves a higher score compared to individual models. The best ensemble model achieves a 76.77\% F1-score. As shown in Table \ref{results-ensemble}, the ensemble method is effective to maximize the performance from a bag of models.

\section{Related work}
Emotional knowledge can be represented in different ways. Word-level emotional representations, inspired from word embeddings, learn a vector for each word, and have shown effectiveness in different emotion related tasks, such as sentiment classification \cite{Tang2016Sentiment}, emotion classification \cite{xu2018emo2vec}, and emotion intensity prediction \cite{park2018plusemo2vec}. Sentence-level emotional representations, such as DeepMoji~\cite{deepmoji}, train a biLSTM model to encode the whole sentence to predict the corresponding emoji of the sentence. 
The learned model achieves state-of-the-art results on eight datasets. 
Sentiment lexicons from \citet{taboada2011lexicon} show that word lexicons annotated with sentiment/emotion labels are effective in sentiment classification. This method is further developed using both supervised and unsupervised approaches in \citet{wang2017sentiment}.  Also,  other models, such as a deep averaging network \cite{iyyer2015deep}, attention-based network \cite{winata2018attention}, and memory network \cite{dou2017capturing}, have been investigated to improve the classification performance. Practically, the application of emotion  classification has been investigated on interactive dialogue systems \cite{bertero2016real,winata2017nora,siddique2017zara,Fung2018EmpatheticDS}.

\section{Conclusion}
In this paper, we compare different pre-trained word embedding features by using Logistic Regression and XGBoost along with flat and hierarchical architectures trained in end-to-end models. We further explore a GP for faster hyper-parameter search. Our experiments show that hierarchical architectures give significant improvements and we further gain accuracy by combining the pre-trained features with end-to-end models.

\bibliography{acl2018}
\bibliographystyle{acl_natbib}

\end{document}